\documentclass[11pt]{article}
\usepackage{fullpage}\usepackage[margin=1.5in]{geometry}

\usepackage{microtype}
\usepackage{subcaption}

\usepackage{natbib}
\renewcommand{\cite}{\citep}

\usepackage{times}
\usepackage{soul}
\usepackage{url}
\usepackage[breaklinks=true]{hyperref}
\usepackage[utf8]{inputenc}
\usepackage{graphicx}
\usepackage{amsmath}
\usepackage{amsthm}
\usepackage{booktabs}
\usepackage{algorithm}
\usepackage{algorithmic}
\usepackage{amsfonts}
\usepackage{textcomp}
\usepackage{xcolor}
\usepackage{enumitem}
\usepackage{multirow}

\usepackage{authblk}

\newcommand{\bfd}{\mathbf{d}}

\newcommand{\bfu}{\mathbf{u}}
\newcommand{\bfe}{\mathbf{e}}

\newcommand{\RecSim}{{\textsc{RecSim}}}
\newcommand{\junk}[1]{}

\title{\RecSim: A Configurable Simulation Platform for Recommender Systems\thanks{\url{https://github.com/google-research/recsim}}}

\author[1]{Eugene Ie$^{\dagger}$}
\author[1]{Chih-wei Hsu}
\author[1]{Martin Mladenov}
\author[1]{Vihan Jain}
\author[2]{\\Sanmit Narvekar$^{\mathsection,}$}
\author[1]{Jing Wang}
\author[1]{Rui Wu}
\author[1]{Craig Boutilier$^{\dagger,}$}
\affil[1]{Google Research}
\affil[2]{Department of Computer Science, University of Texas at Austin}

\begin{document}

\maketitle

\let\oldthefootnote\thefootnote
\footnotetext[1]{Corresponding authors: \url{{eugeneie,cboutilier}@google.com}.}
\footnotetext[2]{Work done while at Google Research.}
\let\thefootnote\oldthefootnote

\begin{abstract}
We propose \RecSim, a configurable platform for authoring simulation environments for recommender systems (RSs) that naturally supports \emph{sequential interaction} with users. \RecSim\ allows the creation of new environments that reflect particular aspects of user behavior and item structure at a level of abstraction well-suited to pushing the limits of current reinforcement learning (RL) and RS techniques in sequential interactive recommendation problems. Environments can be easily configured that vary assumptions about: user preferences and item familiarity; user latent state and its dynamics; and choice models and other user response behavior. We outline how \RecSim\ offers value to RL and RS researchers and practitioners, and how it can serve as a vehicle for academic-industrial collaboration.
\end{abstract}

\section{Introduction}

Practical recommender systems (RSs) are rapidly evolving, as advances in artificial intelligence, machine learning, natural language understanding, automated speech recognition and voice user interfaces facilitate the development of \emph{collaborative interactive recommenders (CIRs)}.
While traditional recommenders, such as those based on collaborative filtering \cite{grouplens:cacm97,breese-cf:uai98,salakhutdinov-mnih:nips07}, typically recommend
items that \emph{myopically} maximize predicted user engagement (e.g., through item rating, score or utility), CIRs explicitly use a
\emph{sequence of interactions} to maximize user engagement or satisfaction.
CIRs often use conversational methods \cite{vinyals:2015:neuralconversation,ghazvininejad:2018:neuralconversation}, example critiquing or preference elicitation \cite{chen:2012:examplecritiquing,christakopoulou:2016:prefelicit}, bandit-based exploration \cite{Li:2010:newsbandits,Li:2016:cfbandits,christakopoulou:2018:banditcir}, or reinforcement learning \cite{sun:2018:convrec} to \emph{explore} the space of options in collaboration with the user to uncover good outcomes or maximize user engagement over extended horizons.

While a topic of increasing research activity in AI---especially in the subareas mentioned above---the deployment of CIRs in practice remains limited. This is due, in no small part, to several challenges that researchers in these areas face when developing modeling techniques and algorithms that adequately reflect qualitative characteristics of \emph{user interaction dynamics}. The importance of modeling the dynamics of user interaction when devising good algorithmic and modeling techniques for CIRs is plainly obvious. The next generation of recommenders will increasingly focus on modeling sequential user interaction and optimizing users' long-term engagement and overall satisfaction.
Setting aside questions of user interface design and natural language interaction,\footnote{We return to these topics in the concluding section.} this makes CIRs a natural setting for the use of \emph{reinforcement learning (RL)}.
Indeed, RSs have recently emerged as a useful application area for the RL community.

Unfortunately, the usual practice of developing recommender algorithms using static data sets---even those with temporal extent, e.g., the MovieLens 1M dataset \cite{harper16:movielens}---does not easily extend to the RL setting involving interaction sequences.
In particular, the inability to easily extract predictions regarding the impact of \emph{counterfactual actions} on user behavior makes applying RL to such datasets challenging. This is further exacerbated by the fact that data generated by RSs that optimize for myopic engagement are unlikely to follow action distributions similar to those of policies striving for long-term user engagement.\footnote{This is itself will generally limit the effectiveness of off-policy techniques like inverse propensity scoring and other forms of importance weighting.}

To facilitate the study of RL algorithms in RSs, we developed \RecSim, a configurable platform for authoring simulation environments to allow both researchers and practitioners to 
challenge and extend existing RL methods in synthetic recommender settings. Our goal is not to create a ``perfect'' simulator; we do not expect policies learned in simulation to be deployed in live systems. Rather, we expect simulations that mirror \emph{specific} 
aspects of user behavior found in real systems to serve as a controlled environment for developing, evaluating and comparing recommender models and algorithms (especially those designed for sequential user-system interaction). As an open-source platform, \RecSim\ will also aid reproducibility and sharing of models within the research community, which in turn, will support increased researcher engagement at the intersection of RL/RecSys. For the RS practitioner interested in applying RL, \RecSim\ can challenge assumptions made in standard RL algorithms in stylized recommender settings, identify pitfalls of those assumptions to allow practitioners to focus on additional abstractions needed in RL algorithms. This in turn reduces live experiment cycle time via rapid development and model refinement in simulation, and minimizes the potential fo negative impact on users in real-world systems.

The remainder of the paper is organized as follows. We provide an overview of \RecSim\ along with its relations with RL and RecSys. We then conclude this introduction by elaborating specific goals (and non-goals) of the platform, and suggesting ways in which both the RecSys and RL research communities, as well as industrial practitioners, might best take advantage of \RecSim. We briefly discuss related efforts in Section~\ref{sec:related}. We outline the basic components of \RecSim\ in Section~\ref{sec:components} and describe the software architecture in Section~\ref{sec:arch}. We describe several case studies in Section~\ref{sec:cases} designed to illustrate some of the uses to which \RecSim\ can be put, and conclude with a discussion of potential future developments in Section~\ref{sec:nextsteps}.

\subsection{\RecSim: A Brief Sketch}

\begin{figure}[t]
\centering
  \includegraphics[width=1\linewidth]{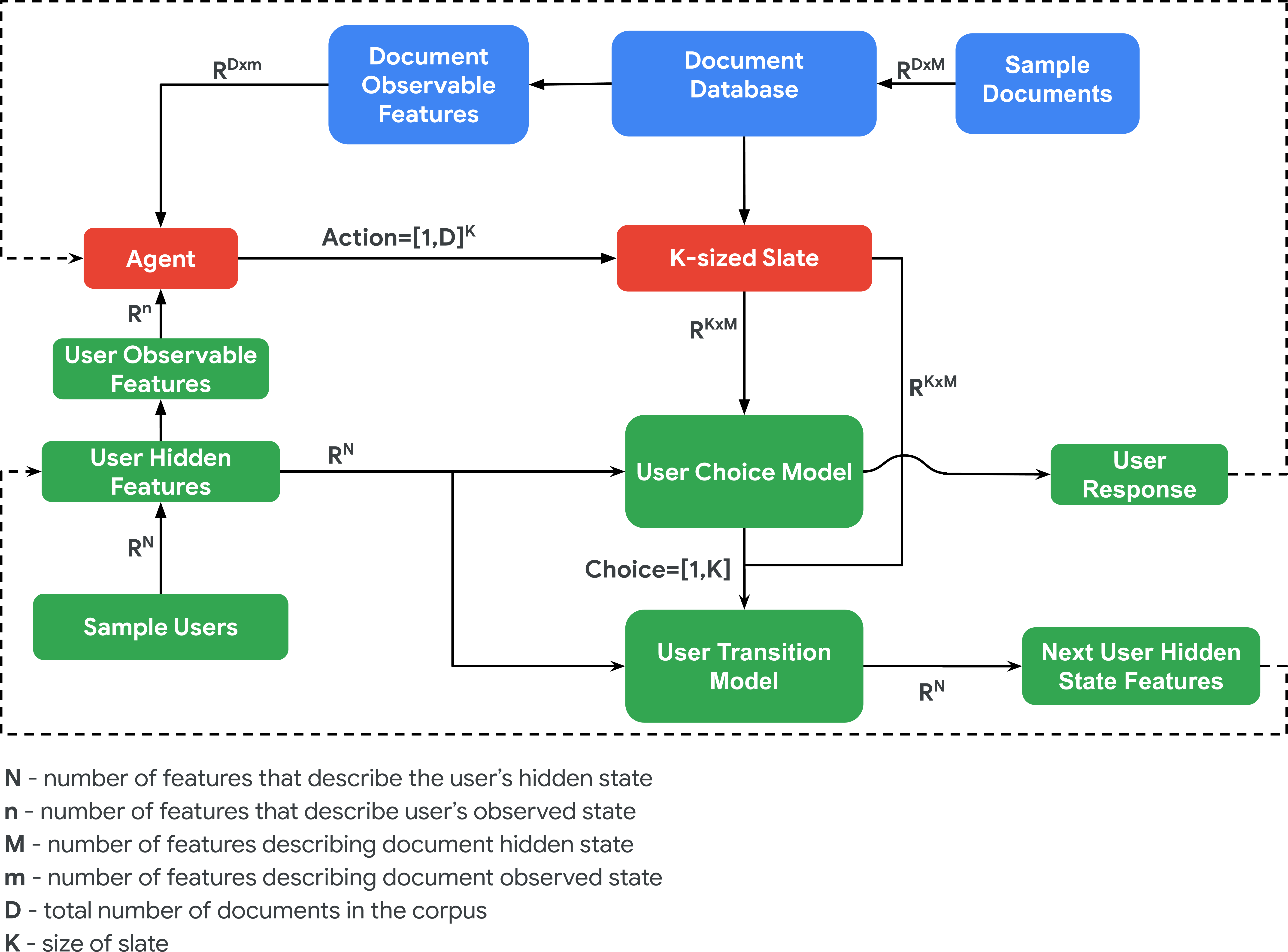}
  \caption{Data Flow through components of \RecSim.}
  \label{fig:system_overview}
\end{figure}

\RecSim\ is a configurable platform that allows the natural, albeit abstract, specification of an environment in which
a recommender interacts with a corpus of \emph{documents} (or recommendable items) and a set of \emph{users}, to support the development of recommendation algorithms. Fig.~\ref{fig:system_overview} illustrates its main components. We describe these in greater detail in Section~\ref{sec:components}, but provide a brief sketch here to allow deeper discussion of our motivations.

The \emph{environment} consists of a \emph{user model}, a \emph{document model} and a \emph{user-choice model}. The \emph{(recommender) agent} interacts with the environment by recommending slates of documents (fixed or dynamic length) to users. The agent has access to observable features of the users and (candidate) documents to make recommendations. The \emph{user model} samples users from a prior distribution over (configurable) \emph{user features}: these may include latent features such as personality, satisfaction, interests; observable features such as demographics; and behavioral features such as session length, visit frequency, or time budget. The \emph{document model} samples items from a prior over \emph{document features}, which again may incorporate latent features such as document quality, and observable features such as topic, document length and global statistics (e.g., ratings, popularity). The level of observability for both user and document features is customizable, so that developers have the flexibility to capture different RS operating regimes to investigate particular research questions.

When the agent recommends documents to a user, the \emph{user response} is determined by a \emph{user choice model}. The choice of document by the user depends on observable document features (e.g., topic, perceived appeal) and all user features (e.g., interests). Other aspects of a user's response (e.g., time spent with a document or post-consumption rating) can also depend on latent document features (e.g., document quality, length). Once a document is consumed, the user state undergoes a transition through a configurable \emph{(user) transition model}. For example, user interest in a document's topic might increase/decrease; user remaining (time) budget may decrease at different rates depending on document quality; and user satisfaction may increase/decrease depending on document-interest match and document quality. Developers can evaluate overall user engagement in the simulated environment to compare policies derived using different RS or RL models and algorithms. We illustrate different configurations with three use cases in Section~\ref{sec:cases}.

\subsection{\RecSim\ and RL}

One motivation for \RecSim\ is to provide environments that facilitate the development of new RL algorithms for recommender applications. While RL has shown considerable success in
games, robotics, physical system control and computational modeling~\cite{mnih2015,silver2016,haarnoja:2018:softacrobotics,lazic_etal:nips18}, large-scale deployment of RL in real-world applications has proven challenging~\cite{dulac-arnold:2019:realworldrl}.
RSs, in particular, have recently emerged as a useful domain for the RL community that could serve to bridge this gap---the ubiquity of RSs in commercial products makes it ripe for demonstrating RL's real-world impact. Unfortunately, the application of RL to the real-world RSs poses many challenges not widely studied in the mainstream RL literature, among them:

\begin{itemize}
    \item \textbf{Generalization across users}: Most RL research focuses on models and algorithms involving a single environment. 
    A typical commercial RS interacts with millions of users---each user is a distinct (and possibly independent) partially observable Markov decision process (POMDP).\footnote{For the purposes of \RecSim\ we treat an RS's interaction with one user as having no impact on the state of another user. We recognize that multi-agent interaction often occurs across users in practice, and that the RS's objectives (e.g., fairness) may induce further dependence in the policies applied to difference users. We ignore such considerations here (though see the concluding section).} However, as in collaborative filtering, contextual bandits, and related models for recommendation, it is critical that the recommender agent generalizes across users, e.g., by modeling the different environments as a \emph{contextual MDP}~\cite{hallak:2015:cmdp}. Large-scale recommenders rarely have enough experience with any single user to make good recommendations without such generalization.

    \item \textbf{Combinatorial action spaces}: Many, if not most, recommenders propose slates of items to users. Slate recommendation
    has been explored in non-sequential settings, capturing point-wise user choice models using non-parametric means \cite{ai2018sigir,seq2slate_full:arxiv18,jiang_slates_cvae:iclr18}. However modeling such
    a combinatorial action space in the context of \emph{sequential} recommendations poses challenges to existing RL algorithms \cite{sunehag2015deep,metz2017discrete}, as the assumptions they make render them ineffective for exploration and generalization in large-scale recommenders.

    \item \textbf{Large, dynamic, stochastic action spaces}: The set of recommendable items is often generated dynamically and stochastically in many large-scale recommenders. For example, a video recommendation engine may operate over a pool of videos that are undergoing constant flux by the minute: injection of fresh content (e.g., latest news), change in content availability (e.g., copyright considerations or user-initiated deletions), surging/declining content popularity, to name a few. This poses an interesting challenge for standard RL techniques as the action space is not fixed \cite{sasmdps:ijcai18,chandak:2019:sasmdp}.

    \item \textbf{Severe partial observability and stochasticity}: Interaction with users means that an RS is operating in a latent-state MDP; hence it must capture various aspects of the user's state (e.g., interests, preferences, satisfaction, activity, mood, etc.) that generally emit very noisy signals via the user's observed behavior. Moreover, exogenous unobservable events further complicate the interpretation of a user's behavior (e.g., if a user turned off a music recommendation, was it because she did not like the recommendation, or did someone ring her doorbell?). Taken together, these factors mean that recommender agents must learn to act in environments that have extremely low signal-to-noise ratios \cite{advamp:ijcai19}.

    \item \textbf{Long-horizons}: There is evidence that some aspects of user latent state evolve very slowly over long horizons. For example, \citet{hohnhold:kdd15} show that ad quality and ad load induce slow but detectable changes in ads effectiveness over periods of months, while \citet{wilhelm:cikm18} show that video recommendation diversification on YouTube induces similarly slow, persistent changes in user engagement. Maximizing long-term user engagement often requires reasoning about MDPs with extremely long horizons, which can be challenging for many current RL methods \cite{advamp:ijcai19}. In display advertising, user responses such as clicks and conversions can happen days after the recommendation~\cite{thompson_delay:nips11,delayed_feedback:kdd14}, which requires agents to model delayed feedback or abrupt changes in reward signals.

    \item \textbf{Other practical challenges}: Other challenges include accurate off-policy estimation in inherently logs-based production environments and costly policy evaluation in live systems. In addition, there are often multiple evaluation criteria for RSs, among which the tradeoff \cite{multiple_obj:recsys12} and the corresponding reward function may not be obvious.
\end{itemize}

Because of these and other challenges, direct application of published RL approaches often fail to perform well or scale \cite{dulac-arnold:2019:realworldrl,slateQ:ijcai19,advamp:ijcai19}. Broadly speaking RL research has often looked past many of these problems, in part because access to suitable data, real-world systems, or simulation environments has been lacking.

\subsection{\RecSim\ and RecSys}

Environments in which the user's state (both observed and latent) can evolve as the user interacts with a recommender pose new challenges not just for RL, but for RSs research as well. As noted above, traditional research in RSs deals with ``static'' users. However in recent years, RS research has increasingly started to explore sequential patterns in user interaction using HMMs, RNNs and related methods \cite{ruining_he:icdm16,hidasi:2016:rnnrecs,ahmed_smola_et_al:wsdm17}. Interest in the application of RL to optimizing these sequences has been rarer \cite{shani:jmlr05} though the recent successes of deep RL have spurred activity in the use of RL for recommendation \cite{facebook_horizon:2018,zheng:2018:drn,chen_etal:2018top,zhao_slateRL:recsys18,slateQ:ijcai19}. However, much of this work has been developed in proprietary RSs, has used specially crafted synthetic user models, or has adapted static data sets to the RL task.

The use of \RecSim\ will allow the more systematic exploration of RL methods in RS research. Moreover, the configurability of \RecSim\ can help support RS research on more static aspects of recommender algorithms. For instance, user transition models can be ``vacuous'' so that the user state never changes. However, \RecSim\ allows the developer to configure the user state (including it's relationship to documents) to be arbitrarily complex, and vary which parts of the state are observable to the recommender itself. In addition, the user choice model allows one to configure various methods by which users choose among recommended items and their induced responses or behaviors. This can be used to rapidly develop and refine novel collaborative filtering methods, contextual bandits algorithms and the like; or simply to test the robustness of existing recommendation schemes to various assumptions about user choice and response behavior.

\subsection{Non-objectives}

The main goal of \RecSim\ is to allow the straightforward specification and sharing of the main environment components involved in simulating the sequential interaction of an RS with a user. It does \emph{not} (directly) provide learning algorithms (e.g., collaborative filtering or reinforcement learning agents) that generate recommendations. Its main aim is to support the development, refinement, analysis and comparisons of such algorithms. That said, \RecSim\ is distributed with several baseline algorithms---both typical algorithms from the literature (e.g., a simple contextual bandit) and some recent RL-based recommender algorithms, as outlined below---to: (a) allow for straightforward ``out-of-the-box'' testing, and (b) serve as exemplars of the APIs for those implementing new recommender agents.

Instead of providing realistic recommender simulations in \RecSim\ that reflect user behavior with full fidelity, we anticipate that new environments will be created by researchers and practitioners that reasonably reflect particular aspects of user behavior at a level of abstraction well-suited to pushing the capabilities of existing modeling techniques and algorithms. \RecSim\ is released with a variety of different user state, transition and choice models, and several corresponding document models.
Several of these correspond to those used in the case studies discussed in Sec.~\ref{sec:cases}. These are included primarily as illustrations of the principles laid out above. Specifically, we do not advocate the use of these as benchmarks (with the exception of researchers interested in the very specific phenomena they study).

While \RecSim\ environments will not reflect the full extent of user behavior in most practical recommender settings, it can serve as a vehicle to facilitate collaboration and identify synergies between academic and industrial researchers. In particular, through the use of ``stylized user models'' that reflect certain aspects of user behavior, industrial researchers can share both qualitative and quantitative observations of user interaction in real-world systems that is detailed enough to meaningfully inform the development of models and algorithms in the research community while not revealing user data nor sensitive industrial practices. We provide an illustrative example of how this might work in one of the case studies outlined below.

\section{Related Work}
\label{sec:related}

We briefly outline a selection of related work on the use of simulation, first in RL, and next in RS and dialogue systems.

\subsection{RL Platforms}

Simulation has played an outsized role in the evaluation of RL methods in recent years. The Arcade Learning Environment \cite{bellemare:jair2013} (or ALE) introduced a now well-known platform for testing algorithms on a suite of Atari 2600 games. Since then numerous RL evaluation benchmarks and environments have been proposed. We only mention a few here to draw contrasts with our goals, and refer to \citet{castro18dopamine} for an overview of related work and the various goals such platforms can play.

The OpenAI Gym \cite{OpenAI_Gym:arxiv16} is one of the most widely used platforms, consisting of a collection of environments (including both discrete, e.g., Atari, and continuous, e.g., Mujoco-based, settings) against which RL algorithms can be benchmarked and compared. Our work shares OpenAI Gym's emphasis on offering environments rather than agents, but differs in that we focus on allowing the authoring of environments to push development of algorithms that handle new domain characteristics rather than benchmarking. Once configured, however, the \RecSim\ environment is wrapped in an OpenAI Gym environment, which, given its popularity, can facilitate RL experimentation and evaluation.
The Dopamine framework \cite{castro18dopamine}, by contrast, provides simple implementations of a variety of (value-based) RL methods that support the rapid development of new algorithms for research purposes. While easily plugged into new environments---\RecSim\ itself is integrated into Dopamine as discussed below---it does not provide support for authoring environments. Other frameworks also provide standard RL algorithms with libraries integrated with OpenAI Gym 
\cite{facebook_horizon:2018,TFAgents}.

ELF \cite{ELF:nips2017} is a platform that allows configuration of real-time strategy games (RTSs) to support the development of new RL methods to overcome the challenges of doing research with commercial games (e.g., by allowing access to internal game state). It allows configuration of some aspects of the game (e.g., action space) and its parameters, sharing \RecSim's motivation of environment configurability. Like \RecSim, it also supports hierarchy and multi-timescale actions. Unlike \RecSim, ELF pays special attention to the support of different training regimes and is designed for fast performance.

\citet{zhang2018:naturalRLbenchmarks} develop a set of ``natural'' RL benchmarks that augment traditional RL tasks with richer image- and video-based state input to ``widen the state space of the MDP,'' seeking to overcome the simplicity of the state space in many simulated RL benchmarks. They share our motivation to press RL algorithms to address new phenomena, but \RecSim\ focuses on supporting configuring basic new structure in state space, action space, observability, system dynamics, agent objectives, etc., and emphasizes the use of stylized models to challenge the fundamental assumptions of many current RL models, algorithms and training paradigms.

\subsection{RecSys and Dialogue Environments}

\citet{rohde2018recogym} propose RecoGym, a stylized RS simulation environment integrated with the RL-based OpenAI Gym. It provides a configurable environment for studying sequential user interaction combining organic navigation with intermittent recommendation (or ads). While RecoGym supports sequential interaction, it does not allow 
user state transitions; instead the focus in on bandit-style feedback---RL/sequentiality is handled within the learning agent (especially exploration, CTR estimation, etc.). It does allow configuration of user response behavior, item/user dimensionality, etc.

The coupling offline of agent training with simulation of user behavior was studied by \citet{schatzmann:2007:agenda}  using a rule-based approach. Similar rule-based environments have also been recently explored to aid evaluation of goal-oriented dialogue agents \cite{wei_airdialogue:emnlp18}, while  the use of learning to enhance rule-based environment
dynamics has been explored in dialogue-based \cite{peng:2018:deepdynaq} and interactive search \cite{liu_etal:ijcai19} systems. More recently,  generative adversarial networks have been
used to generate virtual users for high-fidelity recommender environments to support
learning policies that can be transferred to real systems \cite{shi_taobao:aaai19,zhao_RLSim:arxiv19}. As environments differ widely across systems and commercial products, we propose that \emph{stylized models} of environments that reflect specific aspects of user behavior will prove valuable in developing new RL/RS approaches of practical import. We thus emphasize ease of authoring environments using stylized models (and, in the future through plugging in learned models), rather than focusing on ``sim-to-real'' transfer using high-fidelity models.
\section{Simulation Components}
\label{sec:components}

We discuss the main components of \RecSim\ in further detail (Sec.~\ref{subsec:main_components}) and illustrate their
role and interaction in a specific recommendation environment (Sec.~\ref{subsec:slateq_components}).

\subsection{Main Components}
\label{subsec:main_components}

Fig.~\ref{fig:system_overview} illustrates the main components of \RecSim. The \emph{environment} consists of a \emph{user model}, a \emph{document model} and a \emph{user-choice model}. The \emph{(recommender) agent} interacts with the environment by recommending slates of documents 
to a user. The agent uses \emph{observable} user and (candidate) document features to make its recommendations. Since observable \emph{history} is used in many RL/RS agents, we provide tools to allow the developer to add various summaries of user history to help with recommendation or exploration.

The \emph{document model} also samples items from a prior over \emph{document features}, including latent features such as document quality; and observable features such as topic, 
or
global statistics (e.g., ratings, popularity). Agents and users can be configured to observe different document features, so developers have the flexibility to capture different RS operating regimes (e.g., model predictions or statistical summaries of other users' engagement with a document may be available to the agent but not observable to the user).

The \emph{user model} samples users from a prior over (configurable) \emph{user features}, including latent features such as personality, satisfaction, interests; observable features such as demographics; and behavioral features such as session length, visit frequency, and (time) budget. The user model also includes a transition model, described below.

 When the agent recommends documents to a user, the \emph{user response} is determined by a \emph{user choice model} \cite{louviere-et-al:statedchoice2000}. The choice of document by the user depends on \emph{observable} document features (e.g., topic, perceived appeal) and all user (latent or observable) features (e.g., interests). Other aspects of the response (e.g., time spent, rating) can themselves depend on \emph{latent} (as well as
 obsevable) document features if desired (e.g., document quality, length). Specific choice models include the multinomial logit \cite{louviere-et-al:statedchoice2000} and exponentiated cascade \cite{joachims2002}. Once a document is consumed, the user state transitions through a configurable \emph{(user) transition model}. For example, user interest in a  document's topic might increase/decrease; user remaining (time) budget may decrease at different rates depending on document quality; and user satisfaction may increase/decrease depending on document-interest match and document quality. Developers can evaluate overall user engagement in the simulated environments to compare policies derived using different RL and recommendation approaches. 

\RecSim\ can be viewed as a dynamic Bayesian network that defines a probability distribution over trajectories 
of slates, choices, and observations. In particular, the probability of a trajectory of user observations $o_t$ and choices $c_t$, recommended slates $A_t$, and candidate documents $D_t$ factorizes as: 
\begin{small}
\begin{align*}
p&(o_1,\ldots,o_N, c_1, \ldots, c_N, A_1,\ldots, A_N)\\
&= \sum_{(z_0,\ldots,z_N)} \Big[p(z_0)p(A_0)p(c_0|A_0, z_0)\\
&\quad\quad\quad\quad\quad\prod_{t=1}^N p(o_t|z_t) p(z_t|z_{t-1}, A_{t}, c_t)p(c_t|A_t, z_{t-1})
 p(A_t|D_t, H_{t-1})p(D_t)\Big],
\end{align*}
\end{small}
where $z_i$ is the user state, $p(z_t|z_{t-1}, A_{t}, c_t)$ the transition model, $p(c_t|A_t, z_{t-1})$ the choice model, $p(o_t|z_t)$ the observation model, and $p(A_t|H_{t-1}, D_t)$ is the recommender policy,
which may depend on the entire history of observables $H_{t-1}$ up to that point.

\RecSim\ ships with several default environments and recommender agents, but developers are encouraged to develop their own environments to stress test recommendation algorithms and approaches to adequately engage users exhibiting a variety of behaviors.

\subsection{SlateQ Simulation Environment}
\label{subsec:slateq_components}

To illustrate how different \RecSim\ components can be configured, we describe a specific slate-based recommendation environment, used by \citet{slateQ:ijcai19}, that is constructed for 
testing RL algorithms with combinatorial actions in
recommendation environments. We briefly review experiments
using that environment in one of our use cases below.

To capture fundamental elements of user interest in the recommendation domain, the environment assumes a set of \emph{topics} (or user interests) $T$. The documents are drawn from content distribution $P_D$ over topic vectors. Each document $d$ in the set of documents $D$ has: an associated \emph{topic vector} $\bfd \in [0,1]^{|T|}$, where $d_j$ is the degree to which $d$ reflects topic $j$; a length $\ell(d)$ (e.g., length of a video, music track or news article); an \emph{inherent quality} $L_d$, representing the topic-independent attractiveness to the average user. Quality varies randomly across documents, with document $d$'s quality distributed according to $\mathcal{N}(\mu_{T(d)},\sigma^2)$, where $\mu_t$ is a \emph{topic-specific} mean quality for any $t\in T$. Other environment realizations may adopt assumptions to simplify the setup, such as, assuming each document $d$ has only a single topic $T(d)$, so $\bfd = \bfe_i$ for some $i\leq |T|$ (i.e., a one-hot topic encoding); 
 using the same constant length $\ell$ for all  documents; or assuming fixed quality variance across all topics.

The \emph{user model} assumes users $u\in U$ have various degrees of interests in topics (with some prior distribution $P_U$), ranging from $-1$ (completely uninterested) to $1$ (fully interested), with each user $u$ associated with an \emph{interest vector} $\bfu\in [-1,1]^{|T|}$. User $u$'s interest in document $d$ is given by the dot product $I(u,d) = \bfu\bfd$. The user's interest in topics evolves over time as they consume different documents. A user's \emph{satisfaction} $S(u,d)$ with a consumed document $d$ is a function $f(I(u,d),L_d)$ of user $u$'s interest and document $d$'s quality. Alternative implementations could include: a convex combination to model user's satisfaction such as $S(u,d) = (1-\alpha) I(u,d) + \alpha L_d$ where $\alpha$  balances user-interest-driven and document-quality-driven satisfaction; or a stylized model that stochastically nudges user interest $I_t$ in topic $t=T(d)$ after consumption of document $d$ using $\Delta_t(I_t) = (-y |I_t| + y)\cdot -I_t$, where $y\in [0,1]$ denotes the fraction of the distance between the current interest level and the maximum level $(1, -1)$. Each user could also be assumed to have a fixed \emph{budget} $B_u$ of time to engage with content during a session. Each document $d$ consumed reduces user $u$'s budget by the document length $\ell(d)$ less a \emph{bonus} $b < \ell(d)$ that increases with the document's appeal $S(u,d)$. Other session-termination mechanisms can also be configured in \RecSim.

To model realistic RSs, the user choice model assumes the recommended document's topic to be observable to the
user before choice and consumption. However, the document's quality is not observable to the user
prior to consumption, but is revealed afterward, and drives the user's state transition.  Popular choice functions like the conditional choice model and exponential cascade model are provided in \RecSim\ to model user choice from a document slate.


\section{Software Architecture}
\label{sec:arch}

In this section, we provide a more detailed description of the simulator architecture and outline
some common elements of the environment and recommender agents.

\subsection{Simulator}

\begin{figure}[t]
\centering
  \includegraphics[width=1\linewidth]{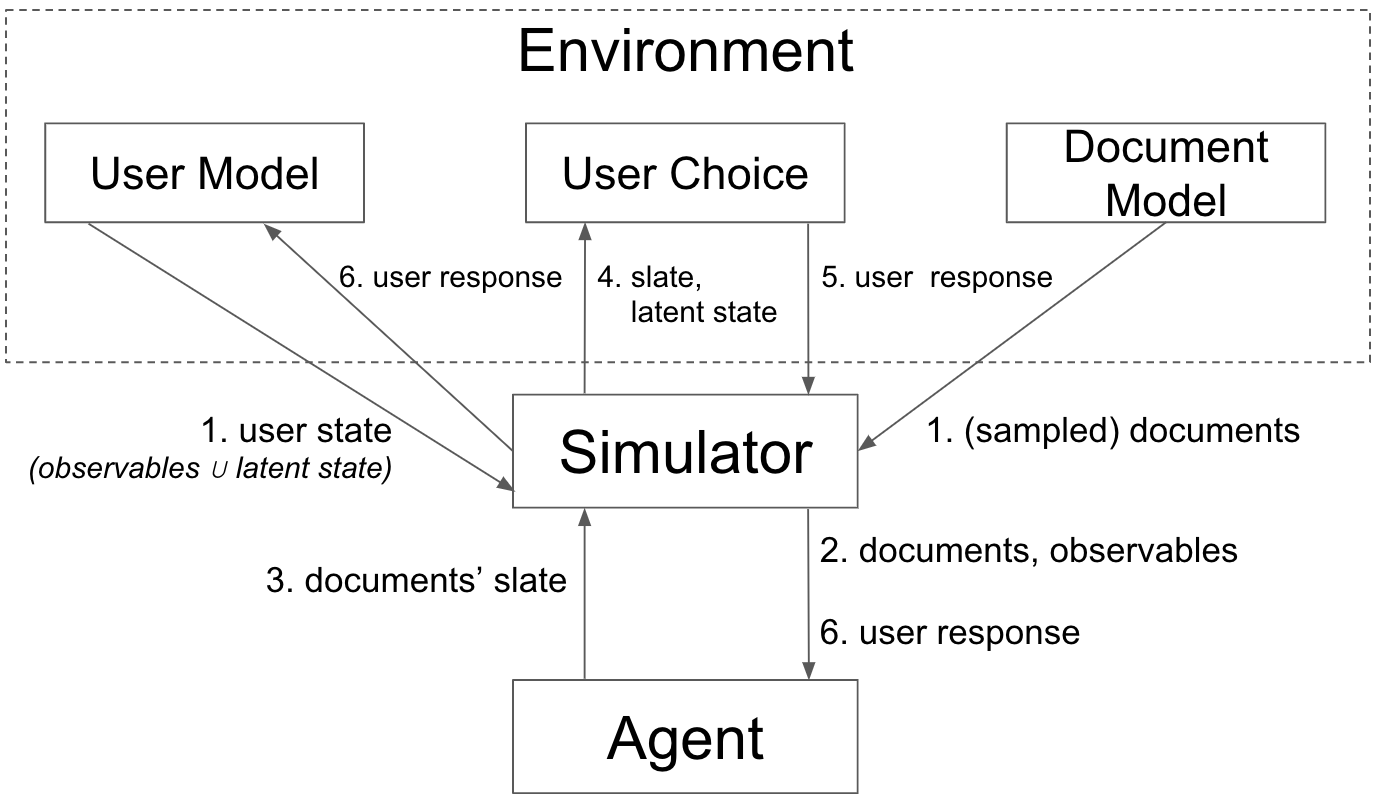}
    \vspace*{-2mm}
  \caption{Control flow (single user) in the RecSim architecture.}
  \label{fig:recsim_architecture}
    \vspace*{-3mm}
\end{figure}

Figure~\ref{fig:recsim_architecture} presents the control flow for a single user in the \RecSim\ architecture. The environment consists of a user model, a document model, and a user-choice model. The simulator serves as the interface between the environment and the agent, and manages the interactions between the two using the following steps:
\begin{enumerate}
\item The simulator requests the user state from the user model, both the observable and latent user features. The simulator also queries the document model for a set of (candidate) documents that have been made available for recommendation. (These documents may be fixed, sampled, or determined in some other fashion.)
\item The simulator sends the candidate documents and the \emph{observable portion} of the user state to the agent. (Recall that the recommender agent does not have direct access to user latent features, though the agent is free to \emph{estimate} them based on its interaction history with that user and/or all other users.)
\item The agent uses its current policy to returns a slate to the simulator to be ``presented'' to the user. (E.g., the agent might rank all candidate documents using some scoring function and return the top $k$.)
\item The simulator forwards the recommended slate of documents and the \emph{full} user state (observable and latent) to the user choice model. (Recall that the user's choice, and other behavioral responses, can depend on all aspects of user state.)
\item Using the specified choice and response functions, the user choice model generates a (possibly stochastic) user choice/response to the recommended slate, which is returned to the simulator.
\item The simulator then sends the user choice and response to both: the user model so it can update
the user state using the 
transition model; and the agent so it can update its policy given the user response to the recommended slate.
\end{enumerate}

In the current design, the simulator sequentially simulates each user.\footnote{We recognize that this design is somewhat limiting. The next planned release of \RecSim\ will support interleaved user interaction.} Each episode is a multi-turn history of the interactions between the agent and single user. At the beginning of each episode, the simulator asks the environment to sample a user model. An episode terminates when it is long enough or the user model transits to a terminal state. Similar to Dopamine \cite{castro18dopamine}, we also define an \emph{iteration},
for bookkeeping purposes, to consist of a fixed number of turns spanning multiple episodes. At each iteration, the simulator aggregates and logs relevant metrics generated since last iteration. The simulator can also checkpoint the agent's model state, so that the agent can restart from that state after interruption.

An important consideration when applying RL to practical recommender applications is the importance of 
\emph{batch RL}
\cite{riedmiller:ecml2005}. A new RS will generally need to learn from data gathered by existing/legacy RSs, since on-policy training and exploration can have a negative impact on users. To facilitate batch RL investigations, \RecSim\ allows one to log the trace of all (simulated) user interactions for offline training using a suitable RL agent. Specifically, it logs each episode as a Tensorflow \cite{tensorflow} \emph{SequenceExample}, against which developers can develop and test batch RL methods (and evaluate the resulting agent through additional environment interaction).

\RecSim\ relies on Tensorflow's Tensorboard to visualize aggregate metrics over time, whether during agent training or at evaluation time using freshly sampled users (and documents if desired). \RecSim\ includes
a number of typical metrics (e.g., average cumulative reward  and episode length) as well as some basic diversity metrics. It uses a separate simulation process to run evaluation (in parallel across multiple users), which logs metrics for a specified number of episodes. The evaluation phase can be configured in multiple ways to evaluate, say, the robustness of a trained recommendation agent to changes in the environment (e.g., changes to user state distribution, transition functions, or choice models). For example, as we discuss in one use case below, we can train the agent using trajectory data assuming a particular user-choice model, but evaluate the agent with simulated users instantiated with a different user-choice model. This separate evaluation process is analogous to the splitting of games into \emph{train} and \emph{test} sets in ALE (to assess over-fitting with hyperperameter tuning).


\subsection{Environment}

The environment provides APIs for the simulator to perform the steps described in Figure~\ref{fig:recsim_architecture}. Once all components in the environment are defined, the environment is wrapped in an OpenAI Gym \cite{OpenAI_Gym:arxiv16} environment. OpenAI Gym has proven to be popular for specifying novel environments to train/test numerous RL algorithms. Developers can readily incorporate state-of-the-art RL algorithms for application recommender domains. Because OpenAI Gym is
intended for RL evaluation, developers are required to define a reward function for each environment---this is usually interpreted as the primary criterion (or at least one of the criteria) for evaluation of the
recommender agent, and will generally be a function of a user's (history of) responses.

%
%
%

\subsection{Recommender Agent Architecture}

\RecSim\ provides several abstractions with APIs for the simulation steps in Figure~\ref{fig:recsim_architecture} so developers can create agents in a configurable way with reusable modules. We do not focus on modules specific to RL algorithms (e.g., replay memory \cite{lin-mitchell}). 
Instead, \RecSim\ offers stackable \emph{hierarchical agent layers} intended to solve a more abstract recommendation problem. A hierarchical agent layer does not materialize a slate of documents (i.e., RS action), but relies on one or more base agents to do so. The hierarchical agent architecture in \RecSim\ can roughly be summarized as follows: a hierarchical agent layer receives an observation and reward from the environment; it preprocesses the raw observation and passes it to one or more base agents. Each base agent outputs either a slate or an abstract action (depending on the use case), which is then post-processed by the agent to create/output the slate (concrete action). Hierarchical layers are recursively stackable in a fashion similar to Keras \cite{keras} layers.

Hierarchical layers are defined by their pre- and post-processing functions and can play many roles depending how these are implemented. For example, a layer can be used as a pure feature injector---it can extract some feature from the (history of) observations and pass it to the base agent, while keeping the post-processing function vacuous. This allows decoupling of feature- and agent-engineering. Various regularizers can be implemented in a similar fashion by modifying the reward. Layers may also be stateful and dynamic, as the pre- or post-processing functions may implement parameter updates or learning mechanisms.    

We demonstrate the use of the hierarchical agent layer using an environment with users having latent topic interests. In this environment, recommender agents are tasked with \emph{exploration} to uncover user interests by showing documents with different topics. Exploration can be powered by, say, a contextual bandit algorithm. A hierarchical agent layer can be used to log bandit feedback (i.e., per-topic impression and click statistics). Its base agent exploits that bandit feedback to returns a slate of recommended documents with the ``best'' topic(s) using some bandit algorithm. These impression and click statistics should not be part of the user model---neither a user's choice nor transition depends on them---indeed, these are agent-specific. However, it is useful to have a common \emph{ClusterClickStatsLayer} for modelling user history in \RecSim\ since many agents often use those statistics in their algorithms. Another set of sufficient statistics that is commonly used in POMDP reinforcement learning is that of finite histories of observables. We have implemented \emph{FixedLengthHistoryLayer} which records observations about user, documents, and user responses during the last few turns. Agents can utilize that layer to, say, model temporal dynamic behavior without requiring direct access to the user model.

A slightly more subtle illustration can be found in the \emph{TemporalAggregationLayer}. It was recently shown~\cite{advamp:ijcai19} how reducing the control frequency of an agent may improve performance in environments where observability is low and/or the available state representations are limited. The \emph{TemporalAggregationLayer} reduces the control frequency of its base layer by calling its \emph{step} function once every $k$ steps, then remembering the features of the returned slate and trying to reproduce a similar slate for the remaining $k-1$ control periods. It also provides an optional switching cost regularizer that penalizes the agent for switching to a slate with different features. In this way, the concept of temporal aggregation/regularization can be applied to any base agent.

Finally, we provide a general hierarchical agent that can wrap an arbitrary list of agents as abstract actions, implementing arbitrary tree-structured hierarchical agent architectures.


\RecSim\ provides a set of baseline agents to facilitate the evaluation of new recommender agents. \emph{TabularQAgent} implements the Q-learning algorithm \cite{watkins:mlj92} (by discretizing observations if necessary). The size of tabular representation of state-action space is exponential in the length of observations; and it enumerates all possible recommendation slates (actions) in order to maximize over Q values during training and serving/evaluation---hence it is a suitable baseline only for the smallest environments. \emph{FullSlateQAgent} implements a deep Q-Network (DQN) agent \cite{mnih2015} by treating each slate as a single action and querying the DQN for maximizing action. The inputs of this DQN are observations of the environment. Slate enumeration generally limits the number of candidate documents that can be evaluated at
each interaction (but see the SlateQ use case below). \emph{RandomAgent} recommends random slates with no duplicates. To be self-contained, \RecSim\ also provides an adapter for applying the DQN agent in Dopamine to the simulation environments packaged with \RecSim. In addition, several standard multi-armed bandit algorithms \cite{auer:2002:ucb1,garivier11klucb,agrawal13further} are provided to support experimentation with exploration  (e.g., of latent topic interests).

\section{Case Studies}
\label{sec:cases}

We outline three use cases developed within \RecSim, one a standard bandit approach, the other two driving recent research into novel RL techniques for recommendation. These illustrate the range of uses to which \RecSim\ can be put, even using environments with fairly simple, stylized models of user interaction.

\subsection{Latent State Bandits} 
In this study, we examine how the tradeoff between immediate and long-term reward affects the value of exploration. The environment samples users whose latent topic interests are not directly observable by the agent. An agent can discover these interests using various exploration strategies. 

This single-item recommendation environment assumes a set of \emph{topics} (or user interests) $T$. The documents are drawn from content distribution $P_D$ over topics. Each document $d$ in the set of documents $D$ has: an associated \emph{topic vector} which is a one-hot topic encoding $\bfd$ of $d$'s sole  topic $T(d)$; an \emph{inherent quality} $L_d$, representing the topic-independent attractiveness. $L_d$ is distributed according to $\ln \mathcal{N}(\mu_{T(d)},\sigma^2)$, where $\mu_t$ is a \emph{topic-specific} mean quality for any $t\in T$. The \emph{user model} assumes users $u\in U$ have various degrees of interests in topics (with some prior distribution $P_U$). Each user $u$ has a static \emph{interest vector} $\bfu$. User $u$'s interest in document $d$ is given by the dot product $I(u, d) = \bfu\bfd$. The probability that $u$ chooses $d$ is proportional to a function depending on topic affinity $I(u, d)$ and document quality: $f(I(u,d) + L_d)$. We evaluate agents using total user
clicks induced over a session.

We can configure $P_U$ and the distribution over $\bfd$ conditioned on $P_U$ so that topic affinity influences
user choice more than document quality: $I(u,d) > L_d$. Intuitively, exploration or planning (RL) is critical in this case---it will be less so when $I(u,d) \sim L_d$ (low topic affinity).
The following table presents the results of applying different exploration/planning strategies,
using the
agents described in Sec.~\ref{sec:arch}: \emph{RandomAgent}, \emph{TabularQAgent}, \emph{FullSlateQAgent}, and a bandit agent powered by UCB1 \cite{auer:2002:ucb1}. The latter three agents employ the \emph{ClusterClickStatsLayer} for per-topic impression and click counts. We also implement an ``omniscient'' greedy agent which knows the user choice model and user prior to myopically optimize expected reward $f(I(\hat{u},d) + L_d)$, where $\hat{u}$ is
an average user (interest). We see that UCB1 and Q-learning perform far better 
than the other agents in the high-affinity environment.

\begin{scriptsize}
\begin{center}
    \begin{tabular}{|@{\hspace{0pt}}r@{\hspace{1pt}}||@{\hspace{1pt}}c@{\hspace{1pt}}|@{\hspace{1pt}}c@{\hspace{1pt}}||@{\hspace{1pt}}c@{\hspace{1pt}}|@{\hspace{1pt}}c@{\hspace{1pt}}|}
    \hline
        Strategy    &   Environment &   Avg.\ CTR (\%) &  
                        Environment &   Avg.\ CTR (\%) \\
    \hline\hline
        Random      &   Low Topic Affinity  &   7.86 &  High Topic Affinity &   14.97\\
        Greedy      &   Low Topic Affinity  &   9.59 (22.01\%) &  High Topic Affinity &   17.56 (17.30\%) \\
        TabularQ    &   Low Topic Affinity  &   8.24 (4.83\%) &  High Topic Affinity &   20.16 (34.67\%) \\
        FullSlateQ  &   Low Topic Affinity  &   9.64 (22.60\%) & High Topic Affinity &   23.28 (55.51\%) \\ 
        UCB1        &   Low Topic Affinity  &   9.76 (24.17\%) &  High Topic Affinity &   25.17 (68.14\%) \\
    \hline
    \end{tabular}
\end{center}
\end{scriptsize}

\subsection{Tractable Decomposition for Slate RL}

Many  RSs recommend \emph{slates}, i.e., multiple items simultaneously,
inducing an RL problem with a large combinatorial action space, that is challenging for exploration, generalization and action optimization. While recent RL methods for such combinatorial action spaces \cite{sunehag2015deep,metz2017discrete} take steps to address this problem, they are unable to scale to problems of the size encountered in large, real-world RSs.

\citet{slateQ:ijcai19}  used \RecSim\ to study decomposition techniques for estimating Q-values of whole recommendation slates. The algorithm exploits certain assumptions about \emph{user choice behavior}---the process by which a user selects and/or engages with items on a slate---to construct a decomposition based on a linear combination of constituent item Q-values. While these assumptions are minimal and seem natural for recommender settings, the authors used \RecSim\ to study (a) the efficacy of the decomposed TD/Q-learning algorithm variants over myopic policies commonly found in commercial recommenders and (b) the robustness of the estimation algorithm under user choice model deviations.

In their simulations, user topic interests are observable and shift with exposure to documents with specific topics. The number of topics is finite, with some having high average quality and others having lower quality. Document quality impacts how quickly a user's (time) budget decays during a session. While users may have initial interest in low-quality topics, their interests shift to higher quality topics if the agent successfully determines which topics have greater long-term value.
Their results demonstrate that using RL to plan long-term interactions can provide significant value in terms of overall engagement. While having full Q-learning that includes optimal slate search in both training and inference may result in the best overall long term user engagement, a significant portion of the gains can be captured using a less costly variant of on-policy TD-learning coupled with greedy slate construction at serving time. They also found that the decomposition technique is robust to user choice model shifts---gains over myopic approaches are still possible even if the assumed user choice model differs. We refer to \citet{slateQ:ijcai19} for additional environment details and results.

\subsection{Advantage Amplification over Long Horizons}

 Experiments in real world ads systems and RSs suggest that some aspects of user latent state evolve very slowly \cite{hohnhold:kdd15,wilhelm:cikm18}. Such slow \emph{user-learning} behavior in environments with a low \emph{signal-to-noise ratio} (SNR) poses severe challenges for end-to-end  \emph{event-level RL}.
 \citet{advamp:ijcai19} used
 \RecSim\ to investigate this issue. The authors developed a simulation environment in which documents have an observable quality that ranges within $[0,1]$. Documents on the $0$-end of the scale are termed \emph{chocolate}, and lead to large amounts of immediate engagement, while documents on the $1$-end, termed \emph{kale}, generate lower engagement, but tend to increase satisfaction. Users' satisfaction is modeled as variable in $[0,1]$ that stochastically (and slowly) increases or decreases with the consumption of different types of content; pure chocolate documents generate engagement drawn from $\ln{\cal N}(\mu_\mathrm{choc}, \sigma_\mathrm{choc})$, pure kale documents resp. from $\ln{\cal N}(\mu_\mathrm{kale}, \sigma_\mathrm{kale})$, while mixed documents interpolate linearly between the parameters of the two distributions in proportion to their kaleness.

 One possible response to the difficulties of learning in slowly evolving environments with low SNR is through the use of temporally-aggregated hierarchical policies. \citet{advamp:ijcai19} implement two approaches---temporal aggregation (repeating actions for some predetermined period $k$ ) and temporal regularization, i.e., subtracting a constant $\lambda$ from the reward whenever $A_t\neq A_{t-1}$ (in terms of document features)---as hierarchical agent layers in \RecSim\ that can modify a base agent. These hierarchical agent nodes amplify the differences between $Q$-values of actions (the advantage function), making the learned policy less susceptible to low-SNR effects. In simulation, temporal aggregation was shown to improve the quality of learned policies to a point almost identical to the case where the user satisfaction is fully observed. We refer to \citet{advamp:ijcai19} for the environment details and results.

\section{Next Steps}
\label{sec:nextsteps}

While \RecSim\ in its current form provides ample opportunity for researchers and practitioners to probe and question assumptions made by RL/RS algorithms in stylized environments, we recognize the broader
interest in the community to develop models that address the ``sim-to-real'' gap. To that end, we are developing methodologies to fit stylized user models using production usage logs---as well as additional hooks in the
framework to plug in such user models---to create environments that are more faithful to specific (e.g., commercial) RSs. We expect that such fitted stylized user models, especially when abstracted suitably to address concerns about data privacy and business practices,
may facilitate industry/academic collaborations. That said, we view this less as directly tackling ``sim-to-real'' transfer, and more as
a means of aligning research objectives around realistic problem characteristics that reflect the needs and behaviors of real users.

Our initial emphasis in this release of \RecSim\ is facilitating the creation of new simulation environments that draw attention to modeling and algorithmic challenges pertinent to RSs. Naturally, there are many directions for further development
of \RecSim. For example, we are extending it to allow concurrent execution that deviates from the single-user control flow depicted in Fig.~\ref{fig:recsim_architecture}. Concurrent execution will not only improve simulation throughput, but also reflects how RS agents operate in real-world production settings. In particular, it will allow investigation of phenomena, such as ''distributed exploration'' across users, that are not feasible within the current serial user control flow.


Finally, modern CIRs will involve rich forms of mixed-mode interactions that cover a variety of system actions (e.g., preference elicitation, providing endorsements, navigation chips) and user responses (e.g., example critiquing, indirect/direct feedback, query refinements), not to mention unstructured natural language interaction. Furthermore, real-world users typically transition across the search-browsing spectrum over multiple RS sessions---our next major release will incorporate some of these interaction modalities. 



\end{document}